\documentclass[conference]{IEEEtran}
\IEEEoverridecommandlockouts
\usepackage{cite}
\usepackage{amsmath,amssymb,amsfonts}
\usepackage{algorithmic}
\usepackage{graphicx}
\usepackage{textcomp}
\usepackage{xcolor}
\usepackage{float}
\usepackage{caption}
\def\BibTeX{{\rm B\kern-.05em{\sc i\kern-.025em b}\kern-.08em
    T\kern-.1667em\lower.7ex\hbox{E}\kern-.125emX}}

\usepackage{balance}
\usepackage{siunitx}
\renewcommand{\baselinestretch}{1.025} 

\usepackage{graphicx}  
\usepackage{multirow}  
\usepackage{comment}
\usepackage{subcaption}
\begin{document}

\title{ViewSparsifier: Killing Redundancy in Multi-View Plant Phenotyping}

\author{
\IEEEauthorblockN{
Robin-Nico Kampa\IEEEauthorrefmark{1},
Fabian Deuser\IEEEauthorrefmark{1},
Konrad Habel\IEEEauthorrefmark{1}, 
and Norbert Oswald\IEEEauthorrefmark{1}
}
\IEEEauthorblockA{\IEEEauthorrefmark{1}%
University of the Bundeswehr Munich, Munich, Germany \\
\{robin-nico.kampa, fabian.deuser, konrad.habel, norbert.oswald\}@unibw.de}
}

\maketitle

\begin{abstract}
Plant phenotyping involves analyzing observable characteristics of plants to better understand their growth, health, and development. In the context of deep learning, this analysis is often approached through single-view classification or regression models. However, these methods often fail to capture all information required for accurate estimation of target phenotypic traits, which can adversely affect plant health assessment and harvest readiness prediction.
To address this, the Growth Modelling (GroMo) Grand Challenge at ACM Multimedia 2025 provides a multi-view dataset featuring multiple plants and two tasks: \textbf{Plant Age Prediction} and \textbf{Leaf Count Estimation}. Each plant is photographed from multiple heights and angles, leading to significant overlap and redundancy in the captured information. To learn view-invariant embeddings, we incorporate 24 views, referred to as the \textbf{selection vector}, in a random selection. Our \textbf{ViewSparsifier} approach won both tasks. For further improvement and as a direction for future research, we also experimented with randomized view selection across all five height levels (120 views total), referred to as \textbf{selection matrices}.
\end{abstract}


\newcolumntype{C}[1]{>{\centering\arraybackslash}p{#1}}

\section{Introduction}
Automated plant phenotyping plays a critical role in addressing global challenges such as food security, climate change adaptation, and sustainable agriculture. In particular, the classification of traits such as leaf count and plant age provides valuable insights for a variety of downstream applications, including nutrient stress detection, early disease diagnosis and pest monitoring \cite{PAUL2025109734,upadhyay2025deep}. These capabilities are critically important for precision agriculture and autonomous farming systems that aim to optimize yield and resource use in real time.
\begin{figure}[t]
  \begin{center}
      \includegraphics[width=\linewidth]{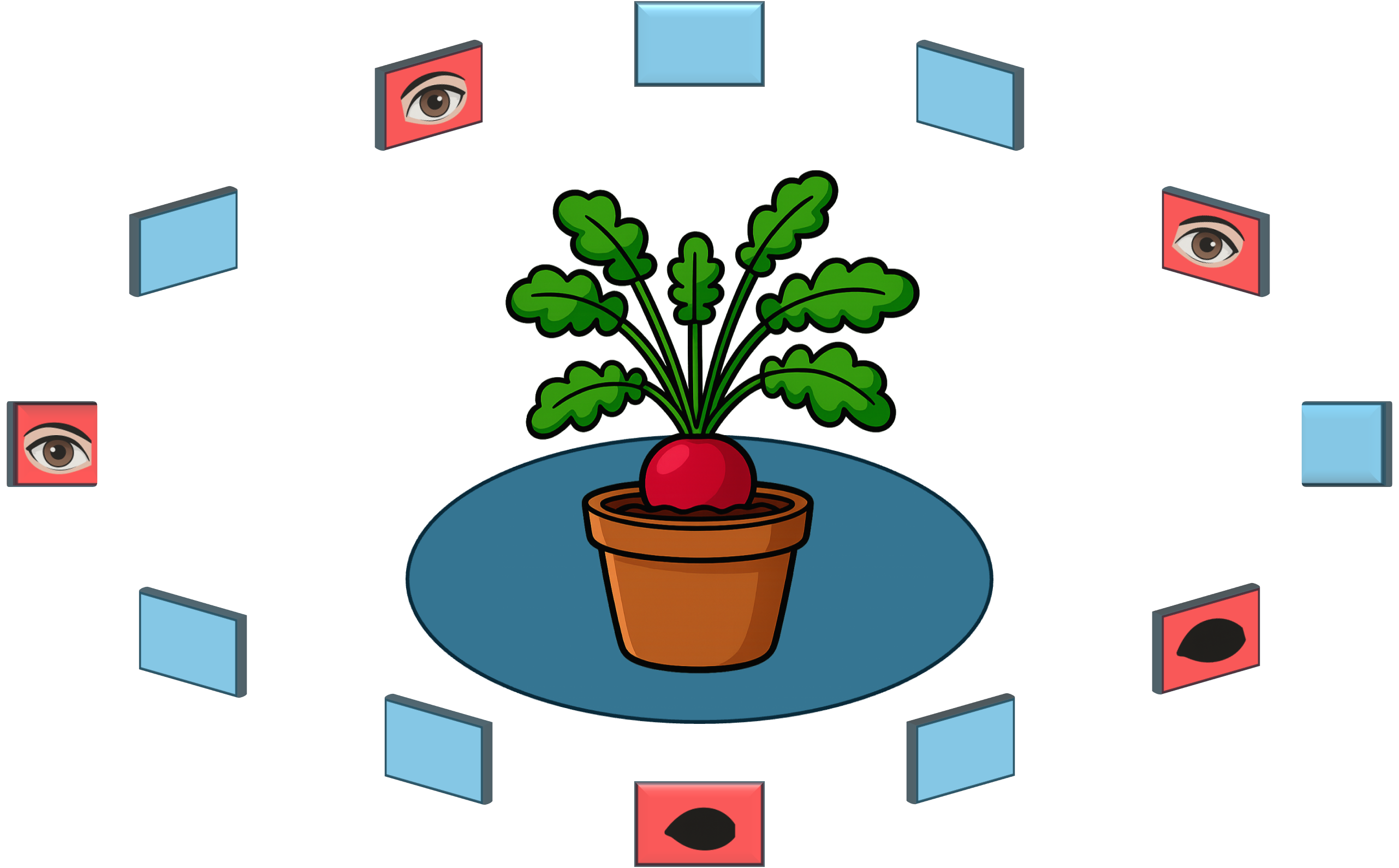}  
  \end{center}
  \caption{Illustration of our \textbf{ViewSparsifier} approach for multi-view plant phenotyping. Each icon represents a possible camera viewpoint around the plant. Red icons mark the actively selected views, while blue icons indicate the remaining unused views. The active views, chosen based on validation performance, form the \textbf{selection vector}. Predictions are averaged across all rotated permutations of the chosen selection to mitigate suboptimal viewpoint selection.
  }
  \label{fig:teaser}
\end{figure}
With the advent of deep learning and high-throughput imaging, it has become possible to quantify plant traits like growth stage and leaf development more efficiently and reproducibly than with traditional methods \cite{murphy2024deep,visionagriculture,KAMILARIS201870}.
According to these works, traditional approaches typically rely on manual measurements, handcrafted features, or classical image processing techniques such as thresholding and simple rule-based edge detection. These approaches are often labor-intensive and tend to lack robustness under real-world field conditions or when scaled to large datasets. Current datasets mostly feature single-image plant phenotyping \cite{minervini2016finely, scharr2014annotated} and suffer from occlusions, with plant parts such as hidden leaves being obscured or located on the opposite side. Multi-view approaches help overcome these limitations by processing multiple views jointly.

\begin{figure}[t]
  \centering
  \includegraphics[width=\linewidth]{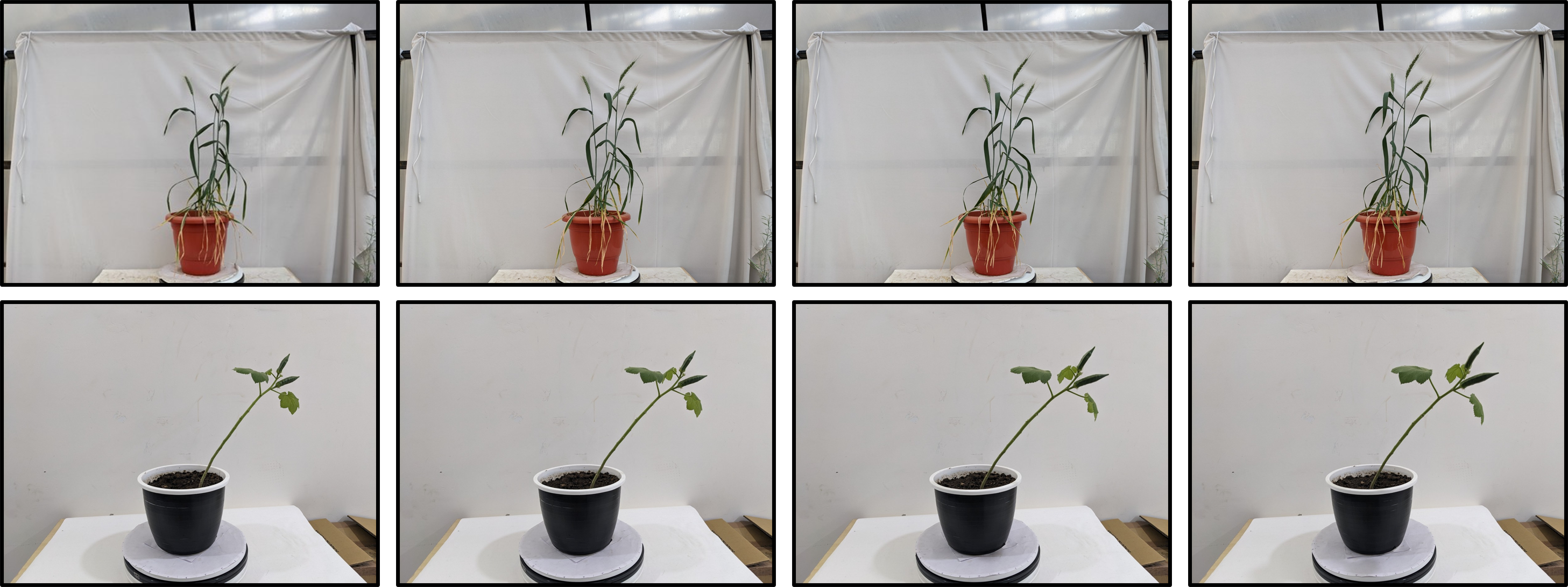}
  \caption{To illustrate the problem of view redundancy: examples from the GroMo~2025 challenge dataset~\cite{bhatt2025gromo} showing a wheat plant (top) at height level~3, day~118, and an okra plant (bottom) at height level~2, day~86. In each case, four directly adjacent views (15° spacing) were captured in sequence, showing only minimal visual differences.}
  \label{fig:gromo-view-redundancy-two-plants}
\end{figure}

The GroMo 2025 Challenge \cite{bhatt2025gromo} contributes to this effort by benchmarking multi-view deep learning methods across multiple crops and traits.
Each plant is captured from five height levels with 15° rotational increments, yielding 24 views per sample.
In this setting, models must infer leaf count and age in days from a large number of slightly varying plant views, which could reflect the future of scalable in-field deployments. Under these conditions, image coverage may be achieved using technologies such as drones. The available information per prediction could then be maximized by using an unlimited number of images captured from all around the plant.
In practice, however, many of these views are visually redundant, as small-degree rotations between consecutive images often result in minimal changes in appearance, as illustrated in Figure~\ref{fig:gromo-view-redundancy-two-plants}, showing examples of a wheat plant and an okra plant from the GroMo 2025 challenge dataset.
This raises a relevant problem for model design, since several consecutive images may appear nearly identical, with new information only emerging after more steps between views. We refer to this phenomenon as view redundancy.\leavevmode\\

Motivated by the problem of view redundancy, we propose ViewSparsifier, a multi-view approach utilizing Transformer-based positional encodings to handle redundancy in multi-view plant phenotyping. The method is specifically designed to cope with large numbers of nearly redundant images.
The main contributions of this work are:\vspace{1.5ex}

\begin{enumerate}
    \setlength{\itemsep}{1ex}
    \item We demonstrate that our view selection strategy, combined with feature aggregation via Transformer-based positional encodings, enables robust predictions despite the high inter-view redundancy present in rotational image sequences.
    \item We propose an extended matrix-like variant that integrates views from multiple height levels using a randomized selection matrix. This approach yields a more comprehensive cross-level representation and achieves improved performance in experimental evaluations.
\end{enumerate}

\section{Related Works}
In recent years, significant progress has been made in automated plant phenotyping through the application of deep learning techniques. Early efforts predominantly relied on convolutional neural networks (CNNs) for single-view classification tasks, such as disease detection and leaf counting \cite{monty, pound2017deep}. However, more recent advancements have shifted towards Transformer-based architectures, as demonstrated in works like \cite{hamdi2024ensemble, chen2023adapting}. Despite the use of CNNs and Vision Transformers (ViTs), these approaches still encountered challenges in generalizing across varying viewpoints, susceptibility to viewpoint bias and occlusion of plant parts.


To address these limitations, multi-view learning has emerged as a promising strategy. Recent studies \cite{daiki2024comparative, wu2022miniaturized, zhang2025wheat3dgs} have introduced advanced strategies for plant phenotyping by leveraging robust 3D reconstruction techniques such as multi-view stereo (MVS) and Gaussian Splatting. For instance, Wu et al. \cite{wu2022miniaturized}
developed a compact MVS-based phenotyping platform tailored for individual plants, while Zhang et al. \cite{zhang2025wheat3dgs} presented Wheat3DGS, a system that combines in-field 3D reconstruction, instance segmentation and wheat head phenotyping using Gaussian Splatting.
In contrast, the direct use of multi-view images for phenotyping without intermediate 3D reconstruction remains relatively underexplored. Nonetheless, new architectures and methods such as the GroMo 2025 baseline \cite{bhatt2025gromo} and the approach by Riera et al. \cite{riera2021deep} showed that effective multi-view phenotyping is possible without explicit 3D modeling.

However, these methods often overlook a key challenge in multi-view learning: redundancy.
Recent work has shown that redundant or irrelevant information can significantly impair model performance \cite{ou2025incomplete, wang2022countering, mao2025investigating}. In the context of multi-view multi-label classification, Ou et al. \cite{ou2025incomplete} highlight the issue of redundant views, where overlapping content from different perspectives introduces noise and inefficiency. Similarly, Wang et al. \cite{wang2022countering} distinguished between irrelevant information (features unrelated to the task) and repetitive information (content repeated across modalities), both of which can obscure important modality-specific signals.
This problem also appears in modern Multimodal Large Language Models (MLLMs), where it manifests as encoder redundancy. Mao et al. \cite{mao2025investigating} showed that using multiple vision encoders, though intended to provide complementary perspectives, can lead to overlapping or even conflicting outputs. This complicates fusion, increases computational cost and may reduce overall performance.
Altogether, these findings highlight redundancy as a pervasive and growing challenge across different domains. Addressing this issue is crucial for developing efficient and robust models, particularly in multi-view plant phenotyping, where input views are abundant yet often redundant. It is essential to avoid overwhelming models with repetitive information and irrelevant image content. 

To our knowledge, no prior work has systematically explored random view selection and permutation-based inference in the context of multi-view plant phenotyping, particularly with the goal of reducing view redundancy and thereby making better use of redundant observational data.
Explicitly targeting redundancy through selection and permutation strategies could improve the efficiency and predictive performance of multi-view phenotyping systems and may also be applicable to other domains facing similar challenges.

\section{Methodology}
\begin{figure}
    \begin{center}
        \includegraphics[width=1\linewidth]{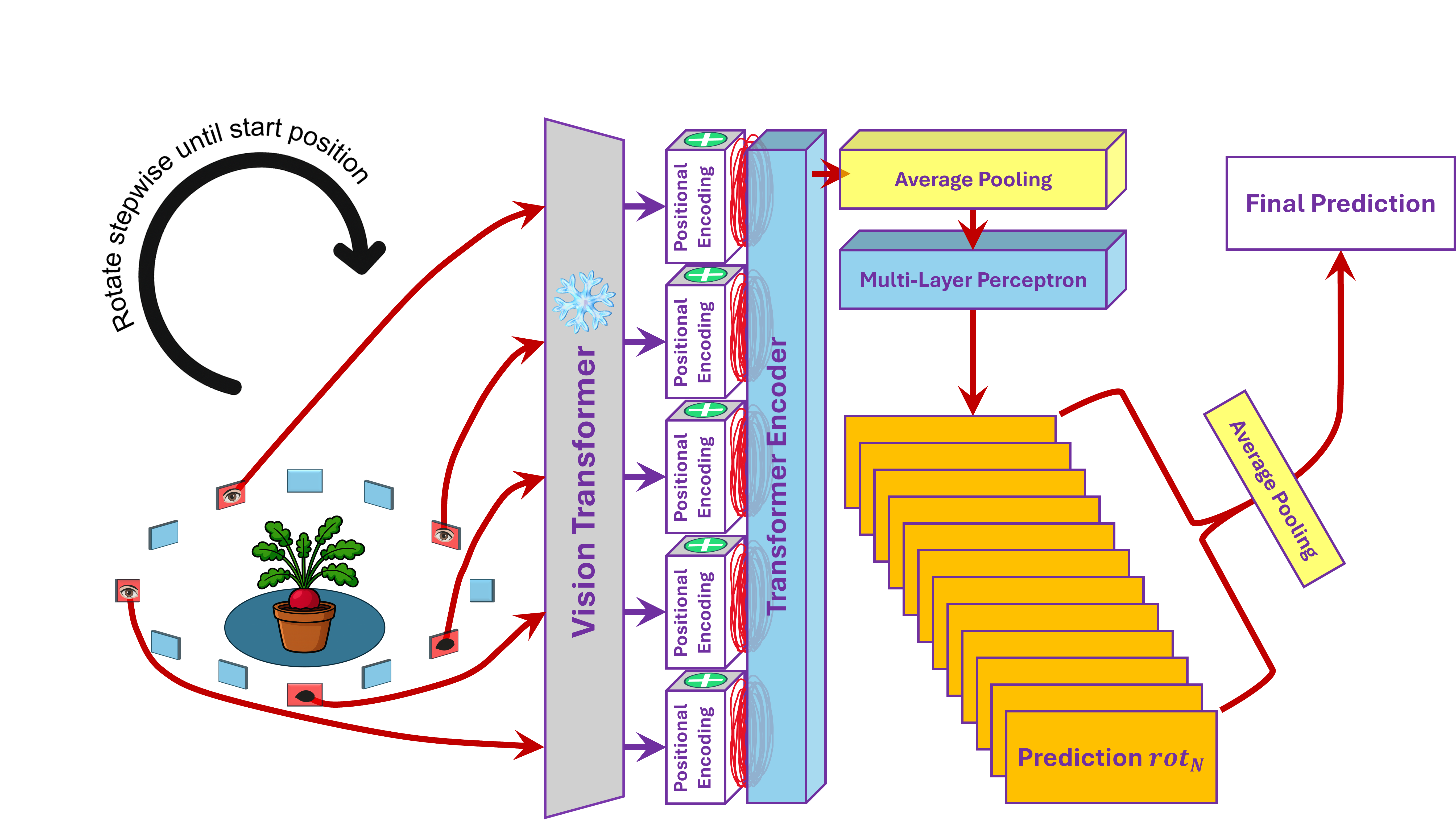}  
    \end{center}
    \caption{The ViewSparsifier architecture: per-view features from a frozen ViT, combined with positional encodings, fused via a Transformer Encoder, and fed into an MLP for regression. During training, a random rotation of the view selection vector is used; during inference, predictions from all rotations are averaged. A non-frozen ViT is applied only for Wheat Age Prediction and Mustard Leaf Count Estimation.}
    \label{fig:architecture}
\end{figure}

As shown in Figure~\ref{fig:architecture}, our ViewSparsifier approach uses features extracted by a frozen Vision Transformer (ViT). The ViT is trained only if doing so proves beneficial, i.e., if it yields measurable improvements in evaluation results. In both cases, whether frozen or unfrozen, we rely on a pre-trained ViT.
The extracted features from the selected views are combined with positional encodings and fused via mean pooling of the Transformer Encoder output, producing a compact representation of the multi-view information. The resulting embedding is then passed to the regression head, implemented as a two-layer MLP with a PReLU activation function~\cite{he2015delving} to introduce non-linearity. This unified architecture is employed for both tasks and across all four crop types of the GroMo 2025 challenge, resulting in a total of eight specialized models. To mitigate overfitting and improve generalization, dropout regularization is applied, with dropout rates individually optimized for each specific crop-task combination.

During training, a single rotational permutation of the selection vector is randomly sampled for each instance to increase data variability and prevent the model from overfitting to a fixed view order. In the GroMo~2025 Challenge, each height level contains 24 distinct views around the plant; therefore, a rotational permutation corresponds to a circular shift of the selected views by anywhere from 1 to 23 positions, effectively altering the viewpoint configuration while preserving the relative ordering of the views.

To increase robustness to suboptimal viewpoint selection, we apply a permutation-based averaging scheme during inference. Each multi-view selection is view-wise rotated around the plant, resulting in 24 rotational permutations. The model processes each rotational permutation of the defined selection vector independently and the predictions are mean-pooled to obtain the final estimate.
To further enhance prediction quality, we remove irrelevant image regions through tailored center cropping (crop-specific, except for wheat), effectively eliminating static, non-informative border areas.

\begin{table*}[ht]
    \centering
    \begin{tabular}{@{}c@{\hskip 2em}c@{}}
        \begin{tabular}{l|
            S[table-format=2.2]
            S[table-format=2.2]
            S[table-format=2.2]
            S[table-format=2.2]|
            S[table-format=2.2]}
            \hline \hline
            \text{Model} & \text{Okra} & \text{Radish} & \text{Mustard} & \text{Wheat} & \text{Mean} \\
            \hline
            Baseline \cite{bhatt2025gromo} & 5.86 & 5.71 & 10.62 & 8.80 & 7.74 \\
            \hline
            CropIQ & 10.80 & 16.54 & 21.70 & 28.60 & 19.41 \\
            Agro\_Geek & 13.42 & 18.85 & 11.30 & 28.45 & 18.01 \\
            Rishi & 12.44 & 10.08 & 12.64 & 16.79 & 12.98 \\
            SoumikDas & 11.14 & 2.68 & 10.18 & 10.60 & 8.65 \\
            PlantPixels & 13.10 & 5.60 &  \phantom{..}\textbf{3.20} & 7.30 & 7.30 \\
            AIgriTech & 3.77 & 5.03 & 8.70 & 8.44 & 6.48 \\
            DeepLeaf & 4.80 & 4.60 & 7.80 & 6.15 & 5.83 \\
            \hline
            Ours$_{\text{t,\hspace*{2.295ex} 15e}}$ & 1.81 & 1.98 & 8.67 & 2.97 & 3.86 \\
            Ours$_{\text{t,\hspace*{2.295ex} 15e, es}}$ & 1.95 & 1.99 & 8.41 & 2.97 & 3.83 \\       
            Ours$_{\text{t+v, 15e}}$ &  \phantom{..}\textbf{1.38} & 2.07 & 7.86 &  \phantom{..}\textbf{2.90} & 3.55 \\
            Ours$_{\text{t+v, 15e, es}}$ &  \phantom{..}\textbf{1.38} & 2.06 & 7.75 &  \phantom{..}\textbf{2.90} & 3.52 \\
            Ours$_{\text{t+v, \phantom{1}4e}}$$^\dagger$ & 1.71 &  \phantom{..}\textbf{1.61} & 6.47 &  \phantom{..}\textbf{2.90} & \phantom{..}\textbf{3.17} \\
            \hline \hline
        \end{tabular}
        &
        \begin{tabular}{l|
            S[table-format=2.2]
            S[table-format=2.2]
            S[table-format=2.2]
            S[table-format=2.2]|
            S[table-format=2.2]}
            \hline \hline
            \text{Model} & \text{Okra} & \text{Radish} & \text{Mustard} & \text{Wheat} & \text{Mean} \\
            \hline
            Baseline \cite{bhatt2025gromo} & 2.04 & 4.04 & 4.99 & 10.80 & 5.52 \\
            \hline
            Agro\_Geek & 1.85 & 3.98 & 7.38 & 30.41 & 10.90 \\
            Rishi & 18.50 & 6.89 & 4.15 & 5.18 & 8.68 \\
            CropIQ & 2.52 & 6.48 & 5.65 & 8.70 & 5.83 \\
            PlantPixels & 2.76 &  \phantom{..}\textbf{0.66} & 8.30 & 5.68 & 4.35 \\
            SoumikDas & 3.06 & 1.33 & 3.33 & 7.20 & 3.73 \\
            DeepLeaf & 0.99 & 0.89 & 7.60 & 5.25 & 3.69 \\
            AIgriTech & 1.65 & 1.79 & 4.40 & 6.71 & 3.63 \\
            \hline
            Ours$_{\text{t+v, 15e}}$ & 1.20 & 0.80 &  \phantom{..}\textbf{2.70} & 3.53 & 2.06 \\
            Ours$_{\text{t,\hspace*{2.295ex} 15e}}$ & 1.02 & 0.80 & 3.00 & 3.40 & 2.06 \\
            Ours$_{\text{t+v, 15e, es}}$ & 1.21 & 0.80 &  \phantom{..}\textbf{2.70} & 3.50 & 2.05 \\
            Ours$_{\text{t,\hspace*{2.295ex} 15e, es}}$ &  \phantom{..}\textbf{0.92} & 0.83 & 3.05 & \phantom{..}\textbf{3.34} & 2.04 \\
            Ours$_{\text{t+v, \phantom{1}4e}}$$^\dagger$ & 1.06 & 0.83 &  \phantom{..}\textbf{2.70} & 3.38 &  \phantom{..}\textbf{1.99} \\
            \hline \hline
        \end{tabular}
        \vspace{1.55ex}\\
        \small (a) GroMo 2025 Challenge Task 1 - Plant Age Prediction & \small (b) GroMo 2025 Challenge Task 2 - Leaf Count Estimation
    \end{tabular}
    \caption{Performance comparison for both tasks in the GroMo 2025 Challenge: Plant Age Prediction and Leaf Count Estimation. Results are reported as Mean Absolute Error (MAE) on the test split, both per crop and as an average. The submissions from the other participating teams are listed under their corresponding team names. $^\dagger$ denotes our best model.\\
    Notation: \texttt{t} = trained on train split, \texttt{t+v} = trained on both train and validation split, X\texttt{e} = X training epochs, \texttt{es} = early stopping.}
    \label{tab:gromo_results}
\end{table*}

\subsection{GroMo 2025 Dataset}
The dataset used in the GroMo 2025 Challenge \cite{bhatt2025gromo} is a multi-view plant phenotyping dataset. Each plant is imaged from 24 different viewpoints at five distinct height levels, resulting in a total of 120 views per plant per day. According to the official challenge rules, each height level must be processed independently, meaning that only the 24 views corresponding to a single height level may be used for its prediction.

The performance is evaluated using Mean Absolute Error (MAE), which is the official challenge metric and Root Mean Square Error (RMSE), which we report additionally for interpretability.

\subsection{Random Vector-like View Selection}
In the GroMo 2025 Challenge, each plant is captured from 24 viewpoints, spaced at 15° intervals. However, there is high redundancy between consecutive images, where even a 30° rotation often results in minimal visual change, as shown in Figure~\ref{fig:gromo-view-redundancy-two-plants}, which makes using all views inefficient. To reduce computational overhead and increase robustness, we explore vector-based view selection strategies.

Initial experiments with fixed-step view skipping (e.g., selecting every second or fourth image) showed inconsistent performance. Instead, we observed that allowing irregular, randomly sampled subsets of views led to better results. To exploit this, we generate random binary selection vectors (of length 24) and evaluate their performance on the validation set. The best-performing selection is then determined per crop and task.

\subsection{Random Matrix-like View Selection}

While the GroMo 2025 Challenge restricts predictions to views from a single height level, we briefly explore the potential of utilizing all available camera height levels (5 in total, each with 24 views). This allows us to extend our previously introduced vector-based view selection strategy into a matrix-based selection strategy, aiming to further improve prediction capabilities.

Instead of selecting views via a binary vector of length 24, we now consider a binary matrix of size $5 \times 24$, where each row corresponds to one height level. Each row encodes a vector of selected views from a particular height level, enabling information-dense combinations of views.
Analogous to the vector-based approach, the model leverages all rotational permutations of the selected views during training and evaluation.
During training, a random rotational permutation of the selection matrix is applied per sample; during evaluation, predictions are averaged over all permutations.
\begin{table*}[t]
\centering
\begin{tabular}{@{}c@{\hskip 2em}c@{}}
%
%
\begin{tabular}{c|p{1.9375cm}|S[table-format=3.0]|C{1.8cm}|C{1.6cm}}
\hline\hline
& \rule{0pt}{2.5ex}Selection & {Views} & $\text{MAE}_{\text{VAL}}$ & $\text{RMSE}_{\text{VAL}}$ \\
\hline
\multirow{5}{*}{\rotatebox{90}{\textbf{Vector}}}
    & all views     & 24 & 0.818 & 1.176 \\
    & every 2nd     & 12 & 0.799 & 1.163 \\
    & every 4th     & 6  & 0.755 & 1.064 \\
    & first view    & 1  & 0.739 & 0.968 \\
    & random        & 6  & \textbf{0.671} & \textbf{0.910} \\
\hline
\multirow{6}{*}{\rotatebox{90}{\textbf{Matrix}}}
    & all views               & 120 & 1.050 & 1.398 \\
    & every 2nd$^{\dagger}$   & 60  & 0.993 & 1.302 \\
    & every 3rd$^{\dagger}$   & 40  & 1.000 & 1.354 \\
    & every 6th$^{\dagger}$   & 20  & 0.793 & 1.086 \\
    & every 12th$^{\ddagger}$ & 10  & 0.899 & 1.188 \\
    & random                  & 5   & \textbf{0.660} & \textbf{0.895} \\
\hline\hline
\end{tabular}
&
%
%
\begin{tabular}{c|p{1.9375cm}|S[table-format=3.0]|C{1.8cm}|C{1.6cm}}
\hline\hline
& \rule{0pt}{2.5ex}Selection & {Views} & $\text{MAE}_{\text{VAL}}$ & $\text{RMSE}_{\text{VAL}}$ \\
\hline
\multirow{5}{*}{\rotatebox{90}{\textbf{Vector}}}
    & all views     & 24 & 1.140 & 1.477 \\
    & every 2nd     & 12 & 1.126 & 1.421 \\
    & every 4th     & 6  & 1.133 & 1.443 \\
    & first view    & 1  & 1.164 & 1.487 \\
    & random        & 11 & \textbf{1.104} & \textbf{1.416} \\
\hline
\multirow{6}{*}{\rotatebox{90}{\textbf{Matrix}}}
    & all views               & 120 & 0.994 & 1.299 \\
    & every 2nd$^{\dagger}$   & 60  & 0.993 & 1.279 \\
    & every 3rd$^{\dagger}$   & 40  & 0.975 & 1.250 \\
    & every 6th$^{\dagger}$   & 20  & 1.042 & 1.345 \\
    & every 12th$^{\ddagger}$ & 10  & 1.016 & 1.308 \\
    & random                  & 17  & \textbf{0.970} & \textbf{1.244} \\
\hline\hline
\end{tabular}
\vspace{1.55ex}\\
\small (a) Radish - Plant Age Prediction Task & \small (b) Wheat - Leaf Count Estimation Task
\end{tabular}
\caption{Comparison of vector- and matrix-based view selection strategies for the Plant Age Prediction and Leaf Count Estimation tasks, evaluated on wheat and radish using the validation split. To ensure a fair comparison, vector-based strategies are evaluated after 15 epochs and matrix-based strategies after 75, as their forward pass processes five height levels simultaneously. $\dagger$ and $\ddagger$ denote circular right shifts of one and two positions between successive rows within the respective selection matrices.}
\label{tab:combined_selection_vertical}
\end{table*}
\subsection{Implementation Details}
\label{sec:impldetails}
We use the C-AdamW optimizer \cite{liang2024cautious} and train in mixed precision for improved computational efficiency. A cosine learning rate scheduler with one warm-up epoch is applied.

In our best-performing configuration, we utilize a DINOv2 Vision Transformer (ViT). The ViT remains frozen (DINOv2-Giant) in all cases except for the Wheat Age Prediction and Mustard Leaf Count Estimation tasks, where a trainable DINOv2-Base model is used. For these two tasks, a base learning rate of $1.0 \times 10^{-5}$ is applied and input images are resized to $518 \times 518$. In all other cases, frozen ViTs extract features from $224 \times 224$ inputs.

%
The fusion module, which implements the Transformer Encoder and Positional Encodings, is trained with a learning rate of $4.2 \times 10^{-5}$, while the regression head is optimized with a higher learning rate of $7.5 \times 10^{-4}$. Unless otherwise specified, all models are trained for a total of 15~epochs, which we found to be sufficient for convergence without overfitting. An exception is made for the Wheat Age Prediction and Mustard Leaf Count Estimation tasks, where a non-frozen ViT is included in the training process. For these two cases, a shortened 4-epoch training setup was used due to improved validation scores.
Our best configurations use task-specific fusion dropout of $0.0$ or $0.1125$ and a batch size of 32. 

All experiments were conducted on an NVIDIA DGX-2 system with 16 Nvidia V100 GPUs each with 32GB VRAM.

\section{Evaluation}
In Table~\ref{tab:gromo_results}, we present the results of the GroMo 2025 Challenge. Our approach achieves the highest mean performance across both tasks,
the weights of the \textbf{DINOv2-Giant} model are frozen for all experiments, except for \textbf{Wheat Age Prediction} and \textbf{Mustard Leaf Count}, where trainable \textbf{DINOv2-Base} models are used due to superior validation performance. Training on the combined train and validation splits yields the strongest overall mean metric across tasks, with early stopping after four epochs. 
In further ablations, we examine the selection of our feature encoders and explore how our proposed matrix approach could enhance performance. We also investigate computational efficiency. These ablations are only done on Radish Age Prediction and Wheat Leaf Count Estimation.

\begin{table}[h]
\centering
\begin{tabular}{l|c|ccc}
\hline \hline
Selection & Views & Allocated & Reserved & Peak \\
\hline
All Views               & 120   & 181.65 MB & 296.00 MB & 230.97 MB \\
Radish Age$^\dagger$    & 5     & 134.75 MB & 246.00 MB & 213.09 MB \\
Wheat Leaf$^\dagger$    & 17    & 138.76 MB & 254.00 MB & 213.60 MB \\
\hline \hline
\end{tabular}
\caption{GPU memory usage of a single forward pass during training for different matrix-based view selection strategies. The $^\dagger$-marked entries correspond to the best-performing randomly generated matrices obtained for each of the two crop-task combinations shown in Table~\ref{tab:combined_selection_vertical}. 
}
\label{tab:gpu_memory_usage}
\end{table}

\begin{table}[t]
\centering
\begin{tabular}{@{}c|l|c|c@{}}
\hline\hline
 & \rule{0pt}{2.5ex}Visual Encoder Backbone & $\text{MAE}_{\text{VAL}}$ & $\text{RMSE}_{\text{VAL}}$ \\
\hline
\multirow{5}{*}{\rotatebox{90}{\textbf{Radish}}} 
    & ViT So150M2 \cite{vit_so150m2}      & 1.323 & 1.599 \\
    & EVA-02 Large \cite{eva02}           & 1.290 & 1.961 \\
    & SigLIP2 Giant \cite{siglip2}        & 1.021 & 1.346 \\
    & Paligemma So400M \cite{paligemma}   & 0.942 & 1.529 \\
    & DINOv2 Giant \cite{dino_v2}         & \textbf{0.660} & \textbf{0.895} \\
\hline
\multirow{5}{*}{\rotatebox{90}{\textbf{Wheat}}}
    & SigLIP2 Giant \cite{siglip2}        & 1.035 & 1.346 \\
    & ViT So150M2 \cite{vit_so150m2}      & 1.257 & 1.540 \\
    & EVA-02 Large \cite{eva02}           & 1.016 & 1.224 \\
    & Paligemma So400M \cite{paligemma}   & 0.977 & 1.298 \\
    & DINOv2 Giant \cite{dino_v2}         & \textbf{0.970} & \textbf{1.244} \\
\hline\hline
\end{tabular}
\caption{Ablation study of visual encoder backbones for the Radish Age Prediction and the Wheat Leaf Count Estimation tasks, using the best-performing task-specific view selection matrix. All models are evaluated after 75 epochs to match the number of forward passes of the vector-based approach, which is trained for 15 epochs in other ablations.}
\label{tab:encoder_ablation_vertical}
\end{table}

\subsection{Feature Encoder}
For our feature encoder we consider the following open-source methods: ViT So150M2~\cite{vit_so150m2}, EVA-02 Large~\cite{eva02}, SigLIP2 Giant~\cite{siglip2}, Paligemma So400M~\cite{paligemma}, DINOv2 Giant~\cite{dino_v2}, which are a selection of current state-of-the-art image encoders. As we can observe in Table~\ref{tab:encoder_ablation_vertical} DINOv2 provides the best performance followed by Paligemma So400M~\cite{paligemma}, which is why we choose DINOv2 in all of our experiments. This ablation study was conducted using our matrix-like view selection and all feature encoder were frozen.
\subsection{Selection Comparison}
To compare the vector-like and matrix-like approaches, refer to Table~\ref{tab:combined_selection_vertical}.
We compare our best randomly generated selection vector to strategies such as using all views, every second view and other structured vector-based subsets. For the matrix-based selections, we start from a full $5 \times 24$ selection matrix - using all height levels and all views - as an upper-bound baseline. We then compare it to systematically reduced configurations, such as selecting every second or fourth view per level. The lowest entry shows the results of the best randomly generated matrix, which achieves the highest validation performance.
Compared to the vector-like approach used during the competition, the matrix-based strategy yields better results by leveraging all height levels per prediction, enabling a more consistent capture of visual plant traits.
Interestingly, using all available views results in lower performance compared to using randomized views. This suggests that random selection encourages the model to develop a stronger understanding of the data and improves its generalization capability.

\subsection{Computational Efficiency}
To evaluate the computational efficiency impact of our matrix-like view selection and the vector-like view selection compared to using all available views, we report GPU memory consumption and runtime benchmarks in Table~\ref{tab:gpu_memory_usage}. The results are obtained using pre-extracted feature of the DINOv2 Giant.

\section{Discussion}
While our approach ranked first in both tasks of the GroMo 2025 Challenge, it did not achieve the top result across all crop-task combinations. In particular, the Radish Leaf Count and Mustard Age Prediction tasks showed slightly lower performance compared to competing methods. These differences may be partially attributed to suboptimal view selection vectors, as performance margins were narrow and validation fluctuations were observed.

In our ablation study, we compare multiple view selection strategies. Interestingly, using all available views during training does not result in the best performance. Several factors may contribute to this. First, training on all views reduces the number of possible training iterations, leading to faster overfitting and limiting the total training duration. Furthermore, our random selection strategy compels the model to process multiple permutations, encouraging it to extract additional information and improving generalization.

\section{Conclusion}
Our winning solution to the GroMo 2025 Challenge demonstrates the effectiveness of optimized \textbf{selection vectors}, choosing a smaller subset from the 24 available views within a given height level for multi-view plant phenotyping. The \textbf{ViewSparsifier} approach achieves state-of-the-art results in both challenge tasks: \textbf{Plant Age Prediction} and \textbf{Leaf Count Estimation}.
By minimizing redundancy and focusing on the most informative perspective combinations, our approach achieves a better
understanding of plant structures, enabling robust and efficient learning.

Beyond the challenge constraints, we extended this method to a \textbf{matrix-based selection} strategy that incorporates views from all five height levels (up to 120 views), capturing complementary vertical information and producing height-invariant, structurally richer representations. Although not part of the official submission, this variant further improved performance and highlights the importance of view selection for generalizable multi-view analysis.  

The challenge setup assumes fixed camera positions and consistent rotational intervals - conditions rarely met in practice. Real-world data, such as drone or handheld footage, often features unknown and variable positions, orientations, and heights, making view alignment non-trivial. In such settings, positional information must be inferred from visual cues like perspective, occlusion, and scale. Extending our \textbf{ViewSparsifier} to learn from such unsystematic multi-view inputs without explicit pose labels would further increase applicability to practical phenotyping and help eliminate costly calibration procedures.

\section{Acknowledgements}
The authors gratefully acknowledge the computing time grant-
by the Institute for Distributed Intelligent Systems and provided on the GPU cluster Monacum One at the University of the Bundeswehr Munich.

\balance

\bibliographystyle{IEEEtran}
\bibliography{base}

\appendix

\section{Online Resources}
Figure~\ref{fig:teaser} includes AI-generated images created on July 15, 2025, using DALL·E 3 \cite{ramesh2022hierarchical} through the ChatGPT-4o interface solely for illustrative purposes. Specifically, the shown radish plant and the stylized eye were generated with DALL·E 3 and embedded into the visualization, whereas the surrounding red and blue shapes themselves are not AI-generated. Figure~\ref{fig:architecture} contains Figure~\ref{fig:teaser} as part of its composition.

\end{document}